\documentclass[pdflatex,sn-basic,iicol]{sn-jnl}

\usepackage{graphicx}%
\usepackage{multirow}%
\usepackage{amsmath,amssymb,amsfonts}%
\usepackage{amsthm}%
\usepackage{mathrsfs}%
\usepackage[title]{appendix}%
\usepackage{xcolor}%
\usepackage{textcomp}%
\usepackage{manyfoot}%
\usepackage{booktabs}%
\usepackage{algorithm}%
\usepackage{algorithmicx}%
\usepackage{algpseudocode}%
\usepackage{listings}%
\usepackage{xcolor,colortbl}

\theoremstyle{thmstyleone}%
\theoremstyle{thmstyletwo}%

\theoremstyle{thmstylethree}%

\begin{document}

\title[RealDiff: Real-world 3D Shape Completion using Self-Supervised Diffusion Models]{RealDiff: Real-world 3D Shape Completion using Self-Supervised Diffusion Models}


\author*[1]{\fnm{Başak Melis} \sur{Öcal}}\email{b.m.ocal@uva.nl}

\author[2]{\fnm{Maxim} \sur{Tatarchenko}}\email{maxim.tatarchenko@de.bosch.com}

\author[1]{\fnm{Sezer} \sur{Karaoğlu}}\email{s.karaoglu@uva.nl}

\author[1]{\fnm{Theo} \sur{Gevers}}\email{th.gevers.uva.nl}

\affil*[1]{\orgdiv{UvA-Bosch Delta Lab}, \orgname{University of Amsterdam}, \orgaddress{\city{Amsterdam}, \postcode{1098 XH}, \country{Netherlands}}}

\affil[2]{\orgdiv{Bosch Center for AI}, \orgname{Robert Bosch GmbH}, \orgaddress{\city{Renningen}, \postcode{71272}, \country{Germany}}}


\abstract{Point cloud completion aims to recover the complete 3D shape of an object from partial observations. While approaches relying on synthetic shape priors achieved promising results in this domain, their applicability and generalizability to real-world data are still limited. To tackle this problem, we propose a self-supervised framework, namely RealDiff, that formulates point cloud completion as a conditional generation problem directly on real-world measurements. To better deal with noisy observations without resorting to training on synthetic data, we leverage additional geometric cues. Specifically, RealDiff simulates a diffusion process at the missing object parts while conditioning the generation on the partial input to address the multimodal nature of the task. We further regularize the training by matching object silhouettes and depth maps, predicted by our method, with the externally estimated ones. Experimental results show that our method consistently outperforms state-of-the-art methods in real-world point cloud completion.}




\maketitle
\begin{figure*}[h]
  \centering
   \includegraphics[width=0.95\linewidth]{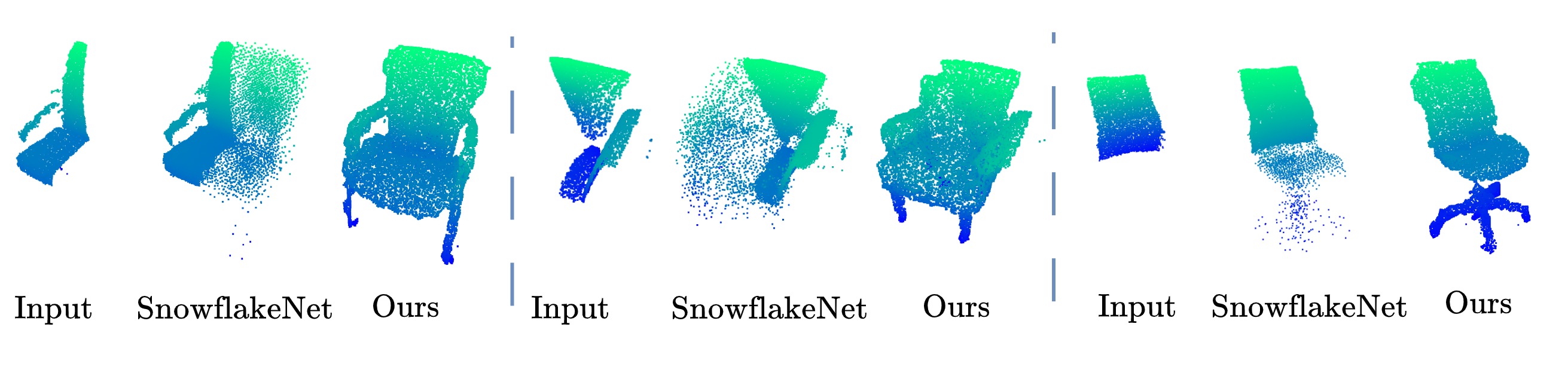}
   \caption{Visual comparison of point cloud completion results. Compared to the SnowflakeNet \citep{snowflakenet} baseline, our method can effectively restore the entire geometry while maintaining the integrity of the original structure.}
   \label{fig:introqual}
\end{figure*}

\section{Introduction}\label{sec:intro}

Driven by the recent advancements in 3D sensor technology, significant strides have been achieved in the field of 3D reconstruction, fostering diverse robotics applications. However, the practical usability of raw point clouds obtained from these contemporary 3D sensors is still frequently compromised by intrinsic limitations, such as self-occlusions, light reflections or limited sensor resolution. These limitations result in sparse and incomplete geometric representations. As a consequence, the task of reconstructing complete point clouds from partial and sparse raw data captured from a single viewpoint is crucial in various robotics applications such as robot navigation, decision-making regarding interactions with objects, and the ability to anticipate the obscured parts of objects.

Some recent point cloud completion methods attempt to tackle the problem by relying on full synthetic supervision \citep{yuan2018pcn, topnet, cascaded, detailpreserved, grnet, pointr, lakenet}, where a parameterized model is usually adopted to learn a mapping from incomplete shapes to complete ones. However, these approaches are not feasible for real-world measurements due to the challenges associated with obtaining paired real-world data. Therefore, prior work resorts to training on paired incomplete shapes, artificially generated from synthetic CAD models, along with their corresponding complete versions. However, this approach falls short in accurately representing the real-world noise, occlusion, and incompleteness patterns, resulting in a prohibitively large synthetic-to-real domain gap. As a promising alternative, unpaired synthetic supervision is employed in more recent developments \citep{pcl2pcl}, where explicit correspondences between partial input data and complete shapes are not required. This is usually achieved by learning a mapping from the latent space of partial shapes to that of complete ones through adversarial training \citep{pcl2pcl, mcn, cycle4} or variational auto-encoding \citep{vrcnet}. Despite their remarkable success, these models, relying on synthetic data to learn the domain of complete shapes, are still unsuitable for real-world applications as the limited set of categories in synthetic benchmarks does not cover the full spectrum of real object categories.

To address the problem of learning 3D shape completion from real-world point clouds acquired by 3D sensors, in this paper, we propose a novel framework called \textit{RealDiff}, aiming to produce high-fidelity and realistic complete object shapes. Our approach is rooted in two key observations: (1) The task of shape completion is inherently multi-modal, i.e. there can be multiple plausible completions of the same partial input. This necessitates the incorporation of probabilistic modeling to address uncertainty in the optimal approach for completing the shapes. In the absence of ground-truth complete data, this probabilistic shape distribution can still be learned through self-supervision by correlating two levels of incompleteness of the same object. (2) Effectively learning shape distributions from noisy and incomplete real-world observations requires additional cues. Leveraging geometric priors offers a way to better handle noisy observations, all without relying on synthetic training data.

To this end, we address the multimodal nature of the problem by adopting a denoising diffusion probabilistic model (DDPM) \citep{diffmodelbase2} as our core module. Given two noisy point clouds of an object captured from two distinctive viewpoints, one of them acts as the input to our pipeline while a pseudo ground-truth is computed by combining the two point clouds. Our method, \textit{RealDiff}, formulates shape completion as a conditional generation problem, by simulating a diffusion process at the missing object parts. To eliminate the noise from the reconstructions, supplementary geometric priors - silhouettes and depth maps - are employed. 
The rendered silhouettes are aligned with the silhouettes obtained by an external method (for instance, ScanNet), and consistency is ensured between the rendered depth maps and those estimated by an off-the-shelf monocular depth estimation network. Hence, the auxiliary silhouettes and depth maps are externally sourced: (1) silhouettes are measured or manually annotated (e.g., ScanNet \citep{dai2017scannet}), and (2) depth is estimated via monocular depth estimation (e.g., \citep{eftekhar2021omnidata}). Incorporating auxiliary weak supervisory signals derived from geometric priors enables the method to achieve enhanced learning of accurate object shapes, as shown in Fig. \ref{fig:introqual}. 

Experiments are conducted with the real-world ScanNet dataset \citep{dai2017scannet} and the synthetic ShapeNet dataset \citep{chang2015shapenet} to verify the merits of our approach. It is shown that the proposed self-supervised method outperforms the state-of-the-art point cloud completion baselines \citep{pointr, adapointr, mcn, shapeformer, p2c}. In brief, our contributions to the existing body of work can be summarized as follows:
\begin{itemize}
    \item We present a novel self-supervised framework for 3D point cloud completion, trained exclusively on real-world observations, without the reliance on synthetic data. To enhance its capability in dealing with imperfect real-world measurements, our framework incorporates additional geometric information such as silhouettes and depth as learning signals. 
    \item Experimental results demonstrate the superiority of the proposed method over state-of-the-art completion methods on real-world point cloud completion task.
\end{itemize}

\section{Related Work}\label{sec:relatedwork}

\subsection{Shape Completion}

3D shape completion is a problem that involves the task of generating a full shape from partial or incomplete input observations. Depending on the scale of the prior information utilized, shape completion methods can be broadly classified into two categories: (1) Surface completion methods relying on local priors and either intend to fill in missing/occluded regions in the surface geometry \citep{sorkine2004least, robustholefilling, laplacianmeshoptimization, kazhdan2006poisson, kazhdan2013screened} or to densify sparse observations \citep{Peng2020ECCV, chibane2020implicit}. (2) Object-level completion methods \citep{pointr, snowflakenet, vrcnet, grnet, pcl2pcl} requiring semantic understanding of the objects to complete partial observations, where certain portions of the inputs are missing. Our work delves into object-level point cloud completion which we briefly review below according to the type of supervision used.

\begin{figure*}[tb]
  \centering
   \includegraphics[width=1.0\linewidth]{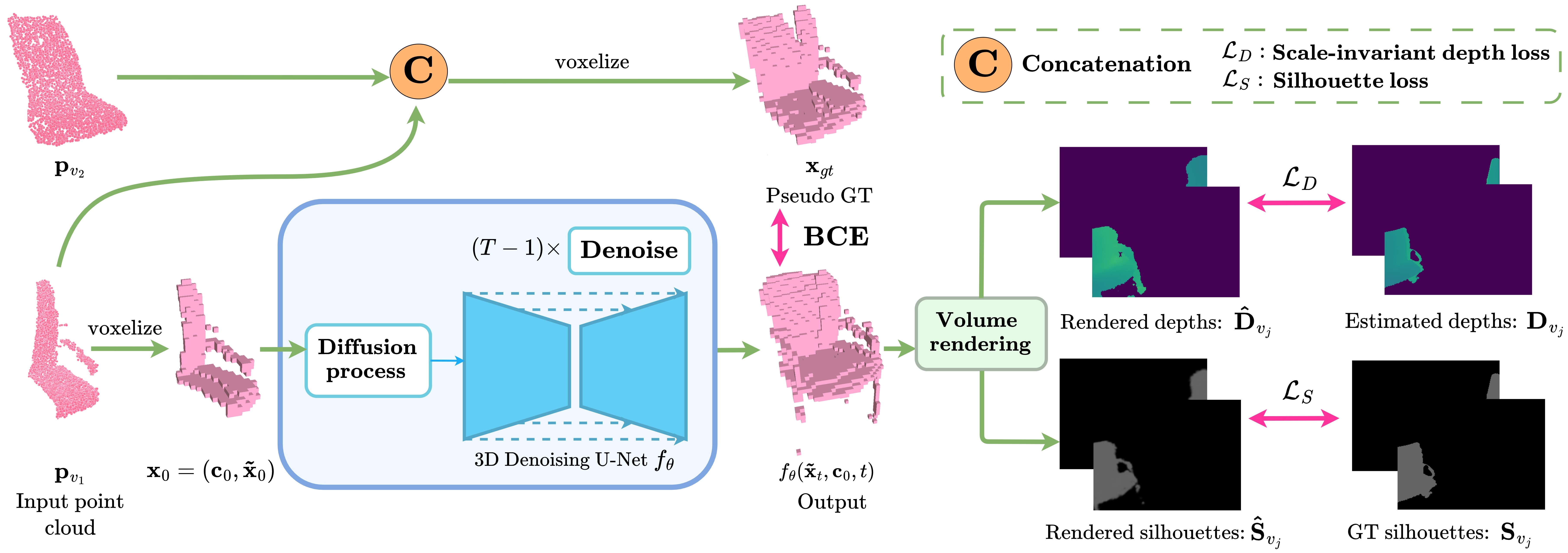}
   \caption{\textbf{Overview of our method.} When given a pair of noisy point clouds representing an object, our pipeline takes one of these point clouds as input, and a pseudo ground-truth is created by combining the two point clouds. A diffusion process is simulated at the missing parts (unoccupied input voxels $\boldsymbol{\mathrm{\tilde{x}}}_{0}$) of the voxelized input $\boldsymbol{\mathrm{x}}_{0}$, while conditioning the generation on the known parts (occupied input voxels $\boldsymbol{\mathrm{c}}_{0}$). To eliminate the noise from the reconstructions, the rendered object shapes’ silhouettes $\boldsymbol{\mathrm{\hat{S}}}_{v_{j}}$ and depth maps $\boldsymbol{\mathrm{\hat{D}}}_{v_{j}}$ are constrained to match the auxiliary silhouettes $\boldsymbol{\mathrm{{S}}}_{v_{j}}$ (e.g. from ScanNet) and depth maps $\boldsymbol{\mathrm{{D}}}_{v_{j}}$ (e.g., from a pre-trained Omnidata model). At generation time, only $f_{\theta}$ is used to reconstruct a complete 3D shape from the real-world point cloud $\boldsymbol{\mathrm{p}}_{v_{1}}$.}
   \label{fig:overview}
\end{figure*}

\noindent\textbf{Full Synthetic Supervision.} Given the partial observation of an object, methods with full synthetic supervision aim to learn a parameterized model as a mapping from incomplete shapes to complete ones. Inspired by point cloud analysis approaches \citep{pointnet, pointnet++}, FoldingNet \citep{yang2018foldingnet} and AtlasNet \citep{atlasnet} are among the first works that directly operate on raw point clouds for object completion. Extending their work, PCN \citep{yuan2018pcn} adopts an encoder-decoder architecture with a refinement phase.
Following this practice, several methods \citep{pfnet, topnet, cascaded, detailpreserved, liu2020morphing} make modifications in their architectures to refine details and to denoise the final predictions. Considering topology information and structures, SA-Net \citep{sa-net}, GRNet \citep{grnet}, PMP-Net \citep{pmpnet} and LAKe-Net \citep{lakenet} utilize detailed geometry information from the input to generate point clouds with increased completeness. Recently, transformer-based methods like PoinTr \citep{pointr} and AdaPoinTr \citep{adapointr} engage in a set-to-set translation between partial scan proxies and the set of occluded proxies, while SnowflakeNet \citep{snowflakenet} models the generation of complete point clouds as a snowflake-like growth of points in 3D space to capture intricate local details. PVD \citep{diffpc1} trains a DDM on point clouds, AutoSDF \citep{autosdf} trains an autoregressive prior, and ShapeFormer \citep{shapeformer} proposes a vector quantized deep implicit function to obtain the sparse representation of the partial point cloud. Similar to our work, DiffComplete \citep{diffcomplete} performs conditional generation using voxel grid diffusion. While it focuses on multi-scale feature interaction of incomplete and complete GT shapes for control, our approach learns shape priors by leveraging geometric priors in a self-supervised way. DiffComplete requires multiple views as input to generate an incomplete 3D scan using volumetric fusion \citep{volumetric}, represented as a truncated signed distance field (TSDF). In contrast, our approach uses only a single view as input, which better reflects real-world scenarios. TSDF approximation with a single view often yields inaccurate surface estimation, making methods relying on TSDF less suitable for practical applications where only a single view is available.
Although all these approaches demonstrate remarkable results, their practicality and adaptability to real-world scenarios are restricted by the difficulties in obtaining paired real-world data. \\
\noindent\textbf{Unpaired Synthetic Supervision} In more recent developments, point cloud completion methods utilizing unpaired synthetic supervision are proposed. Pcl2pcl \citep{pcl2pcl} proposes employing unpaired data and adversarial training to learn a mapping from the latent space of partial shapes to that of complete ones. As a follow-up work, \citep{mcn} introduces the first multimodal shape completion method that completes the partial shapes via conditional generative modeling using unpaired data. Cycle4completion \citep{cycle4} suggests to learn a bidirectional geometric mapping, and thus proposes two simultaneous cycle transformations between the latent spaces. Another work \citep{ganinversion} employs GAN inversion \citep{gainversionpre} exploring a latent code within the pre-trained generator's latent space to identify a complete shape that optimally reconstructs the given input. Although these models have shown success, they still rely on synthetic data to learn the domain of complete shapes, making them unsuitable for real-world observations. Unlike prior work, \cite{weaklysup} also train on real-world, but require more than two views of the objects during training. \cite{p2c} learn completion via patch-wise self-supervised learning, however they do not take into account the multimodal nature of the completion task.

\subsection{DDPMs for 3D Generation and Completion}

Since the seminal works on denoising diffusion probabilistic models (DDPMs) \citep{diffmodelbase1, diffmodelbase2}, several methods extend upon DDPMs by focusing on different sampling strategies \citep{diffsampling1, diffsampling2}, improving efficiency \citep{diffeff1}, and exploring alternative training algorithms \citep{diffextra1, diffextra2, diffextra3}. In the 3D domain, numerous methods explore the generation and completion of point clouds \citep{diffpc1, diffpc2, diffpc3, diffpcextra1}, radiance fields \citep{diffrf1}, and signed distance fields (SDFs) \citep{diffsdf1, diffsdf2, diffsdf3}. However, most of these methods formulate 3D completion as an inpainting task or employ artificially generated incomplete inputs, which does not accurately represent real-world characteristics. Unlike prior work, our approach first voxelizes the point clouds and subsequently employs a 3D U-Net for processing to resolve the issue of unordered point sets.

\section{Method}
\label{sec:methodology}

Given a noisy and partial real-world point cloud of an object, the aim of our novel approach is to complete its 3D shape by predicting the geometry of the unobserved parts. The completion task is formulated as a conditional generation problem that produces the complete shape given the input partial point cloud. The scheme of the pipeline is illustrated in Fig. \ref{fig:overview}. Since the problem of shape completion is multimodal by its nature, a denoising diffusion probabilistic model (DDPM) is used as the core of our approach (Sec. \ref{subsec:subsec1}). Our approach enables the generation of multiple plausible completions from a single incomplete point cloud, while effectively learning category-specific shape priors solely from partial real-world data without considering complete shapes during the training process in a self-supervised way. To overcome the limitations of working with noisy observations without resorting to training on synthetic data, additional geometric cues are used including depth and silhouette information (Sec. \ref{subsec:subsec2}).

\subsection{Learning Shape Representations from Point Clouds}
\label{subsec:subsec1}
Consider two noisy point clouds ${\boldsymbol{\mathrm{p}}_{v_{1}}} \in \mathbb{R}^{N\times3}$ and $\boldsymbol{\mathrm{p}}_{v_{2}} \in \mathbb{R}^{M\times3}$ captured from two different viewpoints of the same object, together with externally obtained silhouette images $\boldsymbol{\mathrm{S}}_{v_{1}}$ and $\boldsymbol{\mathrm{S}}_{v_{2}}$.
In practice these two point clouds can be obtained from any range sensor, for example by back-projecting two depth maps. $\boldsymbol{\mathrm{p}}_{v_{1}}$ denotes the input to our pipeline. Since complete ground-truth shapes are not available, pseudo ground-truth is computed by combining the two point clouds $ \boldsymbol{\mathrm{p}}_{gt} = (\boldsymbol{\mathrm{p}}_{v_{1}},\boldsymbol{\mathrm{p}}_{v_{2}}) \in \mathbb{R}^{(N+M)\times3}$. By using the pseudo ground-truth as the supervisory signal, we correlate two levels of incompleteness of the same object in a self-supervised way. To simplify the 3D data processing with diffusion models, $\boldsymbol{\mathrm{p}}_{v_{1}}$ and $\boldsymbol{\mathrm{p}}_{gt}$ are voxelized into input $\boldsymbol{\mathrm{x}}_{0} = (\boldsymbol{\mathrm{c}}_{0}, \boldsymbol{\mathrm{\tilde{x}}}_{0})$ and (pseudo) ground-truth $\boldsymbol{\mathrm{x}}_{gt}$ occupancy grids, where a voxel is considered occupied if the number of points within it meets or exceeds the voxelization threshold $K$. $\boldsymbol{\mathrm{c}}_{0}$ represents occupied voxels which serve as the condition, while $\boldsymbol{\mathrm{\tilde{x}}}_{0}$ denotes unoccupied voxels. We employ a probabilistic diffusion model and diffuse only the missing shape part $\boldsymbol{\mathrm{\tilde{x}}}_{0}$.

The \textit{forward process} of a DDPM is a Markov chain, which gradually adds Gaussian noise to corrupt $\boldsymbol{\mathrm{\tilde{x}}}_{0}$ into a standard Gaussian distribution $\boldsymbol{\mathrm{\tilde{x}}}_{T}$ in $T$ time steps according to a variance schedule $\beta_{1},....,\beta_{T}$. Then the Markov chain and the Gaussian transition probabilities at each time step is formulated as \citep{diffmodelbase1, diffmodelbase2}:

\begin{equation}
  q(\boldsymbol{\mathrm{\tilde{x}}}_{0:T}) = q(\boldsymbol{\mathrm{\tilde{x}}}_{0}) \prod_{t=1}^{T} q(\boldsymbol{\mathrm{\tilde{x}}}_{t} | \boldsymbol{\mathrm{\tilde{x}}}_{t-1}),
  \label{eq:forward1}
\end{equation}

\begin{equation}
   q(\boldsymbol{\mathrm{\tilde{x}}}_{t} | \boldsymbol{\mathrm{\tilde{x}}}_{t-1}) = \mathcal{N}(\sqrt[]{1-\beta_{t}}\boldsymbol{\mathrm{\tilde{x}}}_{t-1},\beta_{t}\boldsymbol{\mathrm{I}}).
  \label{eq:forward2}
\end{equation}

\noindent Here, the \textit{forward process} is independent of the conditioning factor $\boldsymbol{\mathrm{c}}_{0}$. Similarly, the \textit{reverse process} is defined as a Markov chain with learned Gaussian transitions which aims to iteratively remove the noise added in the \textit{forward process}. The reverse process conditioned on $\boldsymbol{\mathrm{c}}_{0}$ is parameterized by:
\begin{equation}
  p_{\theta}(\boldsymbol{\mathrm{\tilde{x}}}_{0:T},\boldsymbol{\mathrm{c}}_{0}) = p(\boldsymbol{\mathrm{\tilde{x}}}_{T}) \prod_{t=1}^{T} p_{\theta}(\boldsymbol{\mathrm{\tilde{x}}}_{t-1} | \boldsymbol{\mathrm{\tilde{x}}}_{t}, \boldsymbol{\mathrm{c}}_{0}),
  \label{eq:reverse1}
\end{equation}

\begin{equation}
   p_{\theta}(\boldsymbol{\mathrm{\tilde{x}}}_{t-1} | \boldsymbol{\mathrm{\tilde{x}}}_{t}, \boldsymbol{\mathrm{c}}_{0}) = \mathcal{N}(\mu_{\theta}(\boldsymbol{\mathrm{\tilde{x}}}_{t},\boldsymbol{\mathrm{c}}_{0}, t), \sigma_{t}^{2}\boldsymbol{\mathrm{I}}),
  \label{eq:reverse2}
\end{equation}

\noindent where $p(\boldsymbol{\mathrm{\tilde{x}}}_{T})$ is the Gaussian prior, $\sigma_{t}^{2}$ is the variance and $\mu_{\theta}(\boldsymbol{\mathrm{\tilde{x}}}_{t},\boldsymbol{\mathrm{c}}_{0}, t)$ is a neural network that estimates the mean of $p_{\theta}(\boldsymbol{\mathrm{\tilde{x}}}_{t-1} | \boldsymbol{\mathrm{\tilde{x}}}_{t}, \boldsymbol{\mathrm{c}}_{0})$.

DDPMs are trained by optimizing the variational upper bound on the negative log-likelihood, and their training objective can be simplified to an $\mathcal{L}_{2}$ loss between the estimated mean and the target mean \citep{diffmodelbase2}. However, \cite{diffmodelbase2} show that instead of predicting the mean, the model can also be trained by predicting the original sample $\boldsymbol{\mathrm{\tilde{x}}}_{0}$. In this case, the aim is to optimize the binary cross-entropy loss between the predicted and ground-truth occupancy probabilities while training our 3D U-Net $f_{\theta} $ parameterized by $\theta$ as follows:
\begin{equation}
   \mathcal{L}_{t} = \mathcal{L}_{BCE}(f_{\theta}(\boldsymbol{\mathrm{\tilde{x}}}_{t}, \boldsymbol{\mathrm{{c}}}_{0},t), \boldsymbol{\mathrm{x}}_{gt}).
  \label{eq:lossdiffusion}
\end{equation}

\noindent More specifically, we input $\boldsymbol{\mathrm{x}}_{t} = (\boldsymbol{\mathrm{\tilde{x}}}_{t}, \boldsymbol{\mathrm{c}}_{0})$ into $f_{\theta}$ to predict the complete shape $\boldsymbol{\mathrm{{x}}}_{gt}$. Since $\boldsymbol{\mathrm{c}}_{0}$ should remain unaffected through the whole process, we mask out 
the regions of $\boldsymbol{\mathrm{c}}_{0}$ from the model output while computing the loss. 

At generation time, once trained, we can progressively sample as follows:

\begin{equation}
    \resizebox{\columnwidth}{!}{$
    \boldsymbol{\mathrm{x}}_{t-1} = \frac{1}{\sqrt{\alpha_{t}}} \left (\boldsymbol{\mathrm{x}}_{t}- \frac{1- \alpha_{t}}{\sqrt{1 - \tilde{\alpha}}_{t}} (\boldsymbol{\mathrm{x}}_{t} - f_{\theta}(\boldsymbol{\mathrm{\tilde{x}}}_{t}, \boldsymbol{\mathrm{{c}}}_{0},t) \right ) + \sqrt{\beta_{t}}\boldsymbol{\mathrm{z}}   $
    }
  \label{eq:lossdiffusion}
\end{equation}

\noindent with $\alpha_{t} = 1 - \beta_{t}$, $\tilde{\alpha}_{t} = \prod_{s=1}^{t}\alpha_{s}$ and $\boldsymbol{\mathrm{z}} \sim \mathcal{N}(0,\boldsymbol{\mathrm{I}})$. Note that, at generation time, we only require $\boldsymbol{\mathrm{p}}_{v_{1}}$ as input to our network, and we replace the regions of the known shape (conditioned region) in the model output with $\boldsymbol{\mathrm{c}}_{0}$ before computing the sample for the previous time step.

\subsection{Learning Geometric Priors}
\label{subsec:subsec2}

Learning the shape distribution from real-world observations poses a significant challenge, primarily due to the noise in the data arising from inaccurate sensor measurements, occlusions, and errors involved in the segmentation of objects from 3D scenes or images. 
Our approach addresses the task of learning directly from noisy observations by incorporating supplementary geometric priors, such as shape silhouettes and depth maps.

To this end, the predicted object shapes' silhouettes are constrained to match the externally measured silhouettes. 3D reconstructions are converted into 2D silhouettes using volumetric rendering. Hence, to render a silhouette pixel, $M$ points are sampled along its ray $\boldsymbol{\mathrm{r}}$. For each of the 3D points ${m}_{i}$, its occupancy value ${\hat{o}_{i}}$ is obtained from the dense occupancy grid via trilinear interpolation:

\begin{equation}
  {\hat{o}_{i}} = \texttt{interp}({m}_{i}, f_{\theta}(\boldsymbol{\mathrm{\tilde{x}}}_{t}, \boldsymbol{\mathrm{{c}}}_{0},t)),
  \label{eq:interpolation}
\end{equation}


\noindent Following NeRF \citep{mildenhall2020nerf}, the silhouette $\boldsymbol{\mathrm{\hat{S}}} (\boldsymbol{\mathrm{r}})$ for the current ray $\boldsymbol{\mathrm{r}}$ is computed by numerical integration:

\begin{equation}
    \boldsymbol{\mathrm{\hat{S}}} (\boldsymbol{\mathrm{r}}) = \sum_{i=1}^{M}T_{i}\hat{o}_{i},
  \label{eq:volumerendering}
\end{equation}

\noindent where $T_{i} = \exp (-\sum_{j=1}^{i-1} \hat{o}_{j} \delta_j)$ is the accumulated transmittance of a sample point along the ray $\boldsymbol{\mathrm{r}}$, and $\delta_{i}$ is the distance between neighbouring points. Unlike the original numerical quadrature equation which uses alpha values obtained from unbounded densities, we employ a slightly changed formulation by using the already available predicted occupancy probabilities ranging from 0 to 1 instead.

Since two views $v_{1}$ and $v_{2}$ are used to form our ground-truth occupancy grid, the silhouettes are also rendered for the same viewpoints. The $\mathcal{L}_{1}$ loss is computed between the rendered silhouettes $\boldsymbol{\mathrm{\hat{S}}}_{v_{j}}$ and the externally measured silhouettes $\boldsymbol{\mathrm{{S}}}_{v_{j}}$ obtained from the instance masks:

 \begin{equation}
    \mathcal{L}_{S} = \frac{1}{V}\sum_{j=1}^{V} \left\| \boldsymbol{\mathrm{\hat{S}}}_{v_{j}} -\boldsymbol{\mathrm{S}}_{v_{j}}\right\|_{1},
  \label{eq:silhouetteloss}
\end{equation}

\noindent where $V$ denotes the number of views used.

In addition, a depth prior is incorporated by adopting an off-the-shelf monocular depth estimation network. Despite the potential inaccuracy of monocular depths, particularly in low-textured regions, they can still serve as a valuable source of smoothness prior to effectively reduce noise from the reconstructions. This is also the main motivation for using estimated depth map and not the captured one directly - depth measurements are not smooth enough to be used as additional supervision. More specifically, the depth maps are first rendered using the predicted occupancy grid as follows:

\begin{equation}
    \boldsymbol{\mathrm{\hat{D}}} (\boldsymbol{\mathrm{r}}) = \sum_{i=1}^{M}T_{i}\hat{o}_{i}{t}_{i},
  \label{eq:volumerendering}
\end{equation}

\noindent where ${t}_{i}$ denotes the distance of the point from the origin of the camera from which the ray $\boldsymbol{\mathrm{r}}$ is projected. Then, a pre-trained Omnidata model \citep{eftekhar2021omnidata} is used to predict the depth maps $\boldsymbol{\mathrm{D}}_{{v}_{j}}$ for the views used. A scale-invariant loss function is defined to enforce the consistency between the rendered depth maps $\boldsymbol{\mathrm{\hat{D}}}_{{v}_{j}}$ and the estimated monocular depth maps $\boldsymbol{\mathrm{D}}_{{v}_{j}}$ as follows:

 \begin{equation}
    \mathcal{L}_{D} = \frac{1}{V}\sum_{j=1}^{V} \left\| (w\boldsymbol{\mathrm{\hat{D}}}_{v_{j}}+q) -\boldsymbol{\mathrm{D}}_{v_{j}}\right\|_{1},
  \label{eq:depthloss}
\end{equation}

\noindent where $w$ and $q$ are the \textit{scale} and \textit{shift} parameters to align $\boldsymbol{\mathrm{\hat{D}}}_{{v}_{j}}$ and  $\boldsymbol{\mathrm{D}}_{{v}_{j}}$. Following \cite{depthscale1} and \cite{depthscale2}, $w$ and $q$ are solved using least-squares optimization.

Intuitively, the two priors are complementary in their action. While the silhouette forces the network to focus more on predicting the correct shape outline, the depth prior ensures that the surface within the object boundaries is smooth.

The final loss function is a weighted combination of the individual loss terms:

\begin{equation}
\mathcal{L} = \lambda_1\mathcal{L}_{t} + \lambda_2\mathcal{L}_{S} + \lambda_3\mathcal{L}_{D}
\end{equation}

\subsection{Implementation Details}

\textit{RealDiff} is implemented in PyTorch \citep{paszke2019pytorch}. The network training is performed on one NVIDIA RTX A6000 GPU, using an Adam optimizer with batch size 16, and an initial learning rate of 0.0001. For the experiments with geometric priors, we initially train our network without the rendering losses for 250 epochs. Subsequently, we proceed with training using the auxiliary losses until convergence, without the masking out of regions associated with $\boldsymbol{\mathrm{c}}_{0}$ from the model output during loss computation. This approach ensures that, while the diffusion model initially learns to complete shapes by conditioning on  $\boldsymbol{\mathrm{c}}_{0}$ in the first phase of training, it also acquires the ability to remove noise from the reconstructions in the second phase. The voxelization threshold $K$ is set to 10 points. For additional network architecture details, we refer to Appendix \ref{secA1}.

\section{Experiments}
\label{sec:experiments}

In this section, we first introduce our experimental setup in terms of the baselines, benchmarks and evaluation metrics. Then, large scale experiments are presented to evaluate our method for point cloud completion. Sec. \ref{subsec:resultsreal} assesses the methods' point cloud completion capabilities when trained on real-world datasets and Sec. \ref{subsec:multimodalres} demonstrates our model's performance on the multimodal completion task. Additionally, the effectiveness of synthetic shape priors is evaluated in terms of real-world performance (Sec. \ref{subsec:resultssyn}). Finally, an ablation study validating the design choices is given in Sec. \ref{subsec:ablation} and complexity analysis is provided in Sec. \ref{subsec:complexity analysis}. For additional experiments and detailed experimental setup, we refer to the Appendix.

\begin{table*}[h]
\caption{Point cloud completion results on the ScanNet dataset \citep{dai2017scannet} at the resolution of 16384 points. Experimental results show that the proposed method outperforms the state-of-the-art baselines.}
\label{tab:realworldresults1}
\centering
\makebox[\textwidth][c]
{\resizebox{1.0\linewidth}{!}{
    \begin{tabular}{l|ll>{\columncolor[gray]{0.9}}ll|ll>{\columncolor[gray]{0.9}}ll|ll>{\columncolor[gray]{0.9}}ll}
    \toprule
    \multicolumn{1}{c|}{\textbf{Method}}  &  \multicolumn{4}{c|}{Chair} & \multicolumn{4}{c|}{Lamp} & \multicolumn{4}{c}{Table} \\
    
    \cmidrule{2-13} & rec.$\uparrow$   & prec.$\uparrow$  & F1$\uparrow$  & EMD$\downarrow$
    & rec.$\uparrow$ & prec.$\uparrow$  & F1$\uparrow$  & EMD$\downarrow$ 
    & rec.$\uparrow$   & prec.$\uparrow$ & F1$\uparrow$ & EMD$\downarrow$  \\
    \hline \hline
    
    PoinTr \citep{pointr} & 0.12 & 0.23 & 0.14 & 67.61 & 0.11 & 0.28 & 0.16 & 57.62 & 0.16 & 0.43 & 0.22 & 83.38  \\
    Snowflake \citep{snowflakenet} & 0.14 & 0.24 & 0.17 & 64.74 & 0.19 & 0.37 & 0.24 & 61.33 & 0.18 & 0.42 & 0.23 & 84.34  \\
    AdaPoinTr \citep{adapointr} & 0.13 & 0.27 & 0.16 & 66.01 & 0.18 & 0.44 & 0.24 & 65.27 & 0.18 & 0.47 & 0.25 & 74.92 \\
    P2C \citep{p2c} & 0.31 & 0.13 & 0.18 & 78.77 & 0.48 & 0.15 & 0.22 & 56.28  &0.24 & 0.15 & 0.17 & 105.14  \\
    MPC \citep{mcn}  & 0.10 & 0.09 & 0.09 & 90.72 &  0.10 & 0.08 & 0.09 & 91.85 & 0.16 & 0.17 & 0.16 & 98.04 \\
    ShapeFormer \citep{shapeformer}  & 0.17 & 0.21 & 0.18 & 64.66 & 0.11 & 0.11 & 0.10 & 79.45 & 0.12 & 0.11 & 0.11 & 152.19 \\
    \hline 
    Ours & 0.23 & 0.21 & \textbf{0.21} & \textbf{61.29} & 0.25 & 0.30 & \textbf{0.26} & \textbf{47.58} & 0.25 & 0.30 & \textbf{0.26} & \textbf{66.15} \\ 
    \bottomrule
    \end{tabular}
}}
\end{table*}

\begin{figure*}[h]
  \centering
   \includegraphics[width=1.0\linewidth]{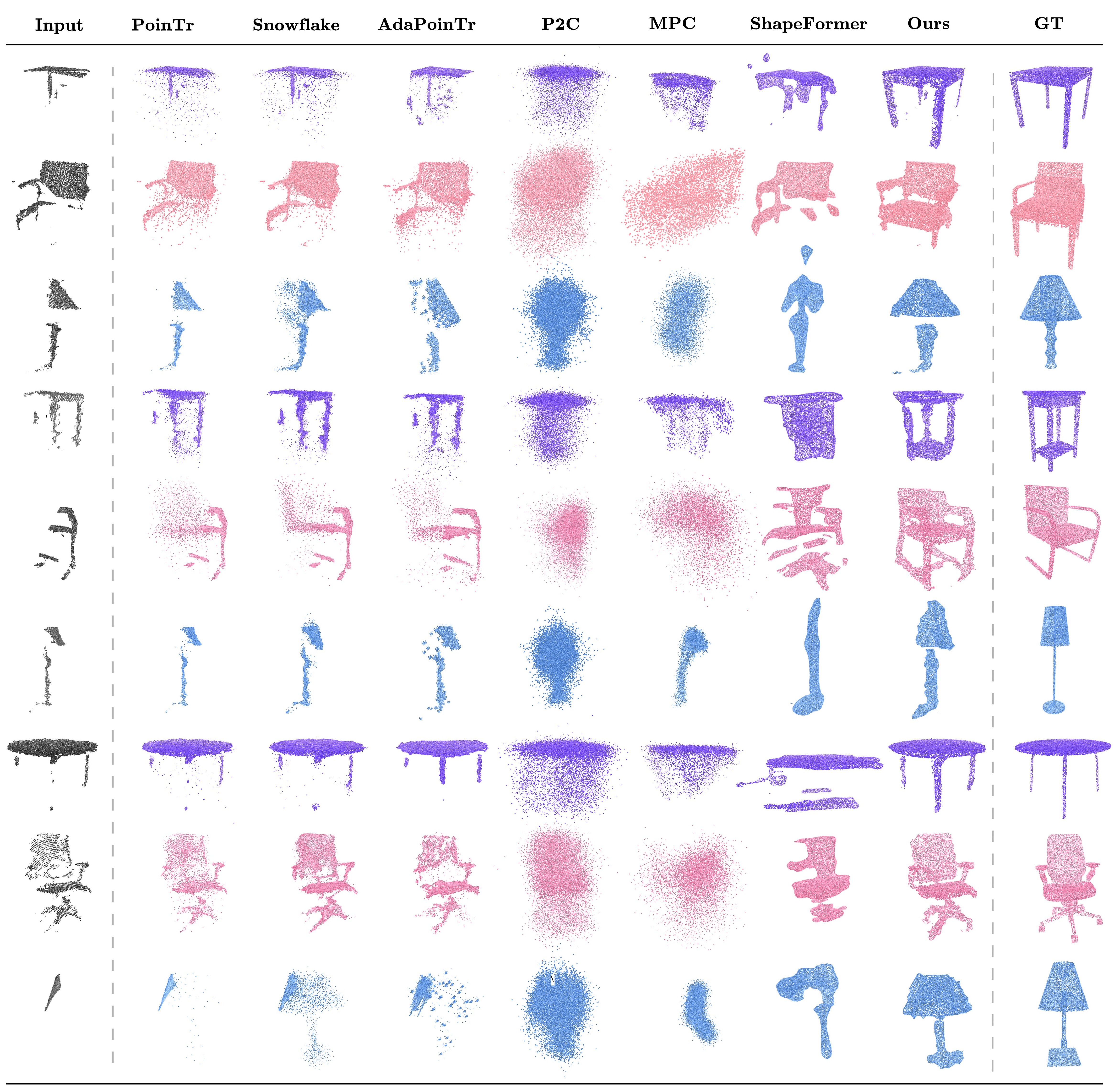}
   \caption{Visual comparison of point cloud completion results on the ScanNet dataset \citep{dai2017scannet}. From left to right: partial shapes sampled from depth images, completion from baselines, our results, and ground-truth CAD model alignments from Scan2CAD annotations. For multi-modal methods, we picked single output shapes corresponding to a specific random seed. Our methodology produces reconstructions that are both more comprehensive and adept at retaining the initially observed structural characteristics.}
   \label{fig:realworldres}
\end{figure*}

\noindent\textbf{Baselines.} Our approach is evaluated and compared with six state-of-the-art point cloud completion methods. PoinTr \citep{pointr}, SnowFlakeNet \citep{snowflakenet} and AdaPoinTr \citep{adapointr} are recent transformer-based methods that generate single completion results. P2C \citep{p2c} is a self-supervised framework that also generates single completion results by grouping incomplete point clouds into local patches and predicting masked patches through learning prior information from different partial objects. MPC \citep{mcn} can produce multiple plausible shapes given the partial input, as it also employs a conditional generative model to learn a mapping from the latent space of partial shapes to that of the complete synthetic ones. ShapeFormer \citep{shapeformer} is also a transformer-based method that proposes probabilistic models for generating multi-modal reconstructions.

\noindent\textbf{Datasets.} A real-world ScanNet dataset \citep{dai2017scannet} and a synthetic ShapeNet dataset \citep{chang2015shapenet} are used. \textbf{ScanNet} is a large-scale dataset of $RGB-D$ scans providing 3D meshes of objects that are pre-segmented from their surrounding environments. Following \cite{mcn}, we focus on three object categories and use $\sim$1500 chairs, $\sim$700 tables and $\sim$110 lamps split into 90\%-10\% train-test sets. All the methods are separately trained and evaluated on these categories for the experiments in real-world. For each object instance, we select at least 8 different views and back-project the corresponding 2.5D images into 3D point clouds. Each point cloud is then sub-sampled to 8192 points. \\
\indent During training, one view is randomly selected as the input. The second view, to form the pseudo ground-truth, is selected from the remaining views based on its contribution to the pseudo ground-truth. To this end, we pick a random view that increases the number of occupied voxels by at least 30\% w.r.t. the first view. The point clouds of these two selected views are combined directly as they share the same coordinate system, and then provided as a pseudo ground-truth with 16384 points to train the methods operating on raw point clouds \citep{pointr, adapointr, snowflakenet, p2c, mcn}. Our method uses the voxelized version of it with $64 \times 64 \times 64$ voxels. ShapeFormer \citep{shapeformer} is provided with both, as it requires both data formats as the input. Externally obtained silhouettes of the objects are used from the readily available per-frame 2D instance masks provided in ScanNet. To evaluate point cloud completion on real-world observations, the Scan2CAD \citep{scan2cad} annotations are used of CAD model alignments from ShapeNet to the extracted objects. For each CAD model, 16384 points are sampled for evaluation. \\
\indent\textbf{ShapeNet} dataset is used to show the effectiveness of synthetic shape priors in real-world performance (Sec. \ref{subsec:resultssyn}). To this end, we also train all the methods on ShapeNet and split the data with the 80\%-20\% strategy. The data pre-processing protocol of \citep{pointr} is taken to obtain input partial observations with 8192 points for training. During evaluation, we randomly select one view from the real-world input data of ScanNet. Predicted outputs with 16384 points are evaluated using CAD models. 

\noindent\textbf{Evaluation Metrics.} For point cloud completion, F1-score is used as the main evaluation metric as suggested by \cite{maximf1}. We additionally report precision and recall to assess \textit{accuracy} and \textit{completeness}. Following prior work \citep{diffpc1, diffpcextra1, ganinversion}, the Earth Mover's Distance (EMD) is used to quantify how uniform the density of the predicted point clouds is (we further refer to this property as \textit{uniformity}). For the probabilistic methods \citep{mcn, shapeformer}, random runs are used for evaluations.

\subsection{Completion Results on Real-world Scans}
\label{subsec:resultsreal}

\noindent\textbf{Results on ScanNet.} We first conduct experiments on the real-world ScanNet dataset \citep{dai2017scannet}. To compare with the existing methods PoinTr \citep{pointr}, SnowflakeNet \citep{snowflakenet}, AdaPoinTr \citep{adapointr}, P2C \citep{p2c}, MPC \citep{mcn} and ShapeFormer \citep{shapeformer}, their open-source code and best hyperparameters from their papers are used for a fair evaluation. Visual comparison of our method with the baselines is provided in Fig. \ref{fig:realworldres}. It can be observed that our approach is able to recover the overall shape of the object together with fine-scale details (\textit{e.g.} chair legs), while also preserving the originally observed structure. On the other hand, transformer-based baselines PoinTr \citep{pointr}, SnowflakeNet \citep{snowflakenet} and AdaPoinTr \citep{adapointr} mostly predict points around the known regions without actually predicting the missing parts. 
The smoothness of our reconstructions surpasses that of the baseline methods, attributable to our utilization of geometric cues to mitigate the noise originating from sensor measurements. In the challenging \textit{lamp} category, all the baselines and our method struggle to recover the details. While our method produces a coarser reconstruction for the primary body of the lamp due to the voxel representation used, it nonetheless accurately recovers the overall shape of the object when compared to the baselines.

Table \ref{tab:realworldresults1} shows the performance of all completion methods. Consistently, our method surpasses the baseline methods when it comes to per-category F1-score and EMD, showcasing the robustness of \textit{RealDiff} in handling various noise and incompleteness patterns encountered in real-world scenarios. 
While our method achieves high recall values across all the categories, it performs worse than the baseline methods in terms of accuracy. Different from the baselines, \textit{RealDiff} employs voxel grid representation to resolve the issue of operating on unordered point sets, which naturally results in offsetting the surface during the final mesh extraction. Also, precision can be highly affected when the networks tend to place more points near the known parts and predict less points in the missing parts as shown in Fig. \ref{fig:realworldres}. Although this strategy can achieve remarkable precision, it frequently culminates in highly deficient reconstructions that do not meet the ultimate goal of the task. Our enhanced recall, on the other hand, more accurately reflects a higher completion capability. Besides, our method reduces the average EMD score by 10.39 when compared against the second-ranked method AdaPoinTr \citep{adapointr} (68.733 in terms of average EMD), enhancing the \textit{uniformity} of the reconstructions. MPC \citep{mcn}, which proposes a conditional generative model to learn a mapping between latent spaces, achieves lower scores. This might be because the latent space of complete shapes (pseudo ground-truths in this case), which are also incomplete, cannot provide a strong supervisory signal when the model is trained on an unpaired setting \citep{mcn}. ShapeFormer \citep{shapeformer} struggles in the real-world scenario, as observations are quantized into patches. Since there are only a few number of patches to extract information from in the case of sparse real-world data, their VQDIF cannot represent the shapes accurately. Similar to ShapeFormer, P2C \citep{p2c} which groups incomplete point clouds into local patches also suffers from the same problem and yields blob-like completion results with lower precision.

\begin{figure*}[h]
  \centering
   \includegraphics[width=0.90\linewidth]{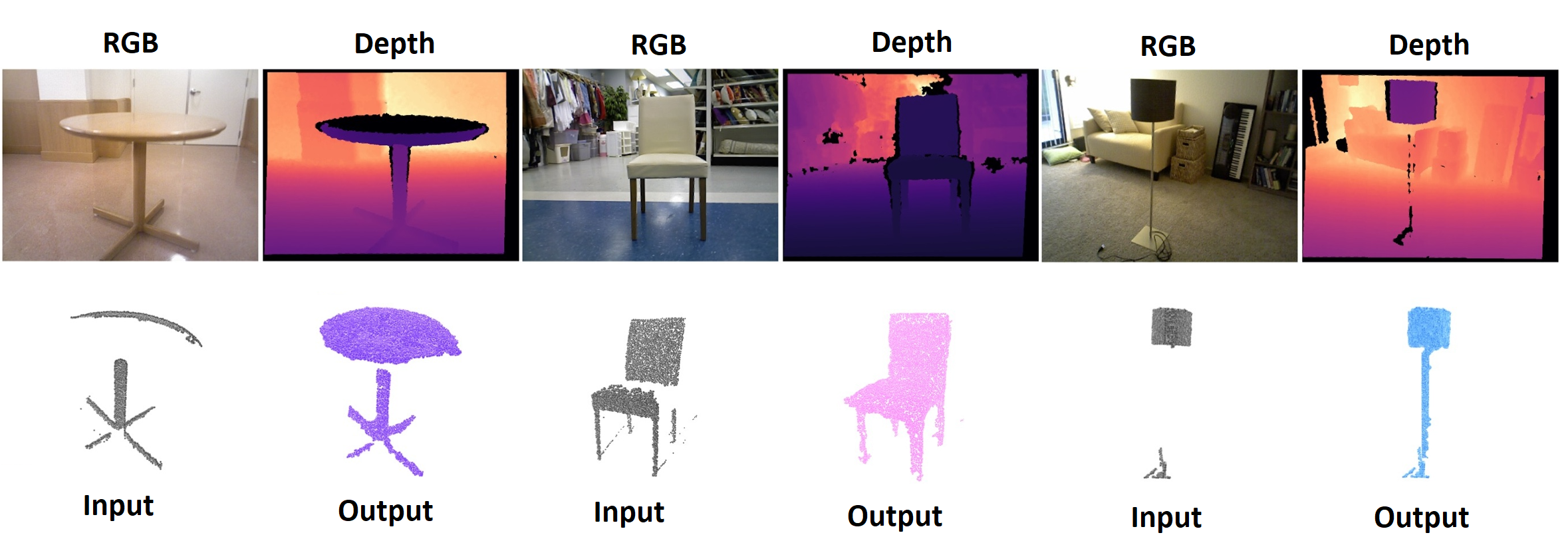}
   \caption{Shape completion on real-world scans from  Redwood 3DScans dataset \citep{redwood3d}. Our pre-trained model on ScanNet \citep{dai2017scannet} is able to generalize to real-world partial shapes from another dataset.}
   \label{fig:redwood}
\end{figure*}

\begin{figure*}[h]
  \centering
   \includegraphics[width=0.9\linewidth]{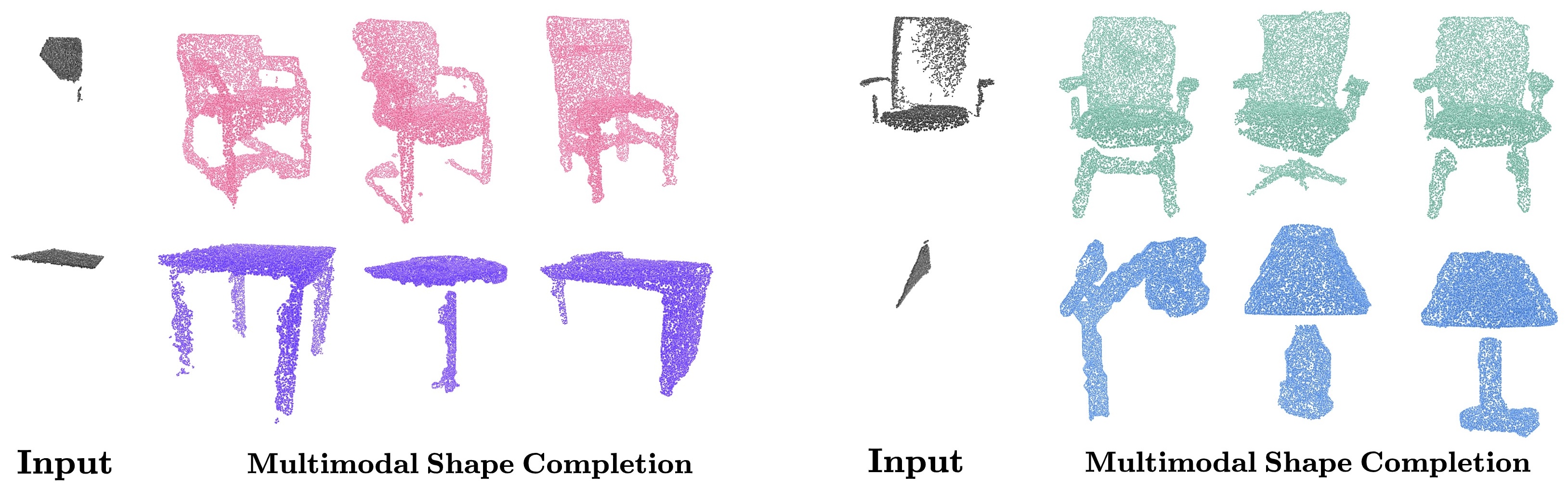}
   \caption{Multimodal completion results for the ScanNet dataset. Shapes are ordered from left to right, top to bottom. Our method is able to generate multiple valid outputs across different runs. As the input incompleteness degree rises, the uncertainty in how to complete the shape geometry also increases which allows for a higher diversity (first, third and fourth shapes). When a more complete input is provided (second shape), we observe slight changes in the recovered geometry between different runs. }
   \label{fig:suppmultimodalqual}
\end{figure*}

\noindent\textbf{Generalization ability and object of interest detection.} ScanNet provides instance segmentation masks, which we utilized to extract the relevant objects. When those are not available, one can rely on off-the-shelf segmentation models. To showcase this and demonstrate the model's ability to generalize across datasets, we apply our ScanNet-pretrained model to scans from the Redwood 3DScans dataset \citep{redwood3d}. We segment the objects of interest using text-conditioned LangSAM (SAM \citep{sam} + GroundingDINO \citep{groundingdino}), and then obtain the input point clouds from the depth images using the segmentation masks. Visual results in Fig. \ref{fig:redwood} show that our pre-trained model is able to generalize to real-world partial shapes from another dataset.

\subsection{Multimodal Completion Results}
\label{subsec:multimodalres}

\textit{RealDiff} addresses the multimodal characteristic of the task by adopting a probabilistic approach, enabling the generation of diverse completion outcomes by employing different Gaussian noise initializations during inference time.

\begin{table}[h]
\caption{Study on the degree of incompleteness of partial inputs on diversity.}
    \centering
    \begin{tabular}{l|lll}
    \toprule
    \multicolumn{1}{c|}{}  &  \multicolumn{1}{c}{1-views} & \multicolumn{1}{c}{5-views} &    \multicolumn{1}{c}{10-views}   \\
    \hline \hline   
    TMD $\uparrow$ & 0.0055  & 0.0052  & 0.0050  \\
    \bottomrule
    \end{tabular}
\label{tab:quantTMDview}
\end{table}

\begin{table*}[h]
\caption{Multi-modal shape completion on ScanNet \citep{dai2017scannet} w/ 16384 points.}
\centering
\resizebox{1.0\linewidth}{!}{
    \begin{tabular}{l|lll|lll|lll}
    \toprule
    \multicolumn{1}{c|}{\textbf{Method}}  &  \multicolumn{3}{c|}{Chair} & \multicolumn{3}{c|}{Lamp} & \multicolumn{3}{c}{Table} \\   
    \cmidrule{2-10} & MMD $\uparrow$ & TMD $\uparrow$ & UHD $\downarrow$  
     & MMD $\uparrow$ & TMD $\uparrow$ & UHD $\downarrow$ & MMD $\uparrow$ & TMD $\uparrow$ & UHD $\downarrow$   \\
    \hline \hline

    MPC \citep{mcn} & 0.110 & \textbf{0.018} & 0.099 & 0.100 & 0.005 & 0.122 & 0.170 & 0.043 & 0.157  \\
    ShapeFormer \citep{shapeformer}   & 0.200 & 0.007 & 0.091 & 0.150 & 0.049 & 0.074 & 0.160 & \textbf{0.168} & 0.163  \\
    \hline 
    Ours & \textbf{0.230} & 0.009 & \textbf{0.087} & \textbf{0.260} & \textbf{0.064} & \textbf{0.060} & \textbf{0.280} & 0.012 & \textbf{0.135}  \\
    \bottomrule
    \end{tabular}}
\label{tab:quantmultimodal}
\end{table*}

\begin{table*}[h]
\caption{Point cloud completion results on the \textit{chair} category of real-world ScanNet dataset \citep{dai2017scannet} when trained on the target category of synthetic ShapeNet dataset \citep{chang2015shapenet}, at the resolution of 16384 points. Synthetic shape priors learned from the same category do not transfer successfully into the real-world.}
\label{tab:syntheticresults}
\centering
\resizebox{0.6\linewidth}{!}{
    \begin{tabular}{l|ll>{\columncolor[gray]{0.9}}ll}
    \toprule
    \multicolumn{1}{c|}{\textbf{Method}}  &  \multicolumn{4}{c}{Chair}  \\
    & rec.$\uparrow$  & prec.$\uparrow$  & F1$\uparrow$ & EMD$\downarrow$ \\
    \hline \hline
    PoinTr \citep{pointr} & 0.140 & 0.224 & 0.166 & 66.73    \\
    Snowflake \citep{snowflakenet} & 0.134 & 0.190 & 0.152 & 69.85  \\
    AdaPoinTr \citep{adapointr} & 0.048 & 0.194 & 0.134 & 68.26 \\
    P2C \citep{p2c} & 0.180 & 0.120 & 0.141 & 88.20
    \\
    MPC \citep{mcn} & 0.187 & 0.132 & 0.151 & 74.34  \\
    ShapeFormer \citep{shapeformer} & 0.186 & 0.161 & 0.169 & 72.54  \\
    \hline
    Ours & 0.197 & 0.163 & 0.174 & 69.78  \\
    \bottomrule
    \end{tabular}}
\end{table*}

Fig. \ref{fig:suppmultimodalqual} shows multimodal shape completion results on ScanNet dataset (Shapes are ordered from left to right, top to bottom). As the degree of incompleteness in the partial input increases, as seen in the first, third, and fourth shapes, the inherent ambiguity leads to a wider range of diverse completions across different runs. Conversely, for the second shape where the input is more complete, there are only slight variations between runs. This is confirmed quantitatively in the following experiment. We provide the model with input point clouds generated from 1, 5, and 10 views, respectively. Then, 10 completed shapes are computed for each input, and the Total Matching Distance (TMD) score is calculated to capture the diversity (higher TMD $\rightarrow$ higher diversity). This yields the results provided in Table \ref{tab:quantTMDview}.

We also perform quantitative evaluation using the metrics suggested by \cite{mcn}: Minimal Matching Distance (MMD) capturing the completion quality, TMD measuring the completion diversity, and Unidirectional Hausdorff Distance (UHD) focusing on the completion fidelity. We provide exact definitions of those metric in Appendix \ref{secA2}. The results are summarized in Table \ref{tab:quantmultimodal}. Our approach achieves the best MMD and UHD scores on all classes, while reaching slightly inferior TMD scores on chairs and tables.

\subsection{Training on Synthetic Data}
\label{subsec:resultssyn}

\noindent\textbf{Synthetic training on the target category.} Prior work employs synthetic priors to learn a mapping between partial and complete shapes. To evaluate the effectiveness of synthetic shape priors from the target category for real-world performance, experiments are conducted on the synthetic ShapeNet dataset \citep{chang2015shapenet}. All methods are trained on the \textit{chair} object category of ShapeNet, and they are directly evaluated on chairs from ScanNet. The quantitative results of baseline methods and ours in Table \ref{tab:syntheticresults} are lower than our method's results provided in Table \ref{tab:realworldresults1}, showing that shape priors learned from synthetic data of the same shape category do not transfer successfully to real-world observations. This observation indicates a substantial domain shift between the synthetic domain and the real-world scenario, underscoring the necessity to train directly on real-world data for the purpose of completing actual observations.

\begin{table*}[h]
\caption{Point cloud completion results on the \textit{chair} category of real-world ScanNet dataset \citep{dai2017scannet} when trained on the other 54 categories of synthetic ShapeNet dataset \citep{chang2015shapenet}, at the resolution of 16384 points. Synthetic shape priors learned from other categories do not transfer successfully to a different target category for the task of real-world shape completion.}
\centering
\resizebox{0.6\linewidth}{!}{
    \begin{tabular}{l|ll>{\columncolor[gray]{0.9}}ll}
    \toprule
    \multicolumn{1}{c|}{\textbf{Method}}  &  \multicolumn{4}{c}{Chair}  \\
    & rec.$\uparrow$  & prec.$\uparrow$  & F1$\uparrow$ & EMD$\downarrow$ \\
    \hline \hline
    PoinTr \citep{pointr} & 0.112 & 0.225 & 0.140 & 67.44    \\
    Snowflake \citep{snowflakenet} & 0.113 & 0.188 & 0.135 & 72.89  \\
    AdaPoinTr \citep{adapointr} & 0.055 & 0.187 & 0.124 & 70.30 \\
    P2C \citep{p2c} & 0.174 & 0.117 & 0.136 & 94.29
    \\
    MPC \citep{mcn} & 0.179 & 0.123 & 0.142 & 78.85  \\
    ShapeFormer \citep{shapeformer} & 0.154 & 0.140 & 0.142 & 89.82 \\
    \hline
    Ours & 0.150 & 0.100 & 0.117 & 107.00  \\
    \bottomrule
    \end{tabular}}
\label{tab:suppsyntheticresults}
\end{table*}

\noindent\textbf{Synthetic training on other categories.} To evaluate the effectiveness of synthetic shape priors from other categories for real-world performance, experiments are conducted on the synthetic ShapeNet dataset \citep{chang2015shapenet}. All methods are trained on the 54 object categories of ShapeNet, except the \textit{chair} category, and they are directly evaluated on chairs from ScanNet. The quantitative results are given in Table \ref{tab:suppsyntheticresults}. When trained on categories other than the target category, the F1-score and the EMD decrease for all the methods compared to the quantitative results provided in Table \ref{tab:syntheticresults}. This indicates that neither synthetic training on the target category, nor synthetic pre-training on large shape quantities from other object categories, is sufficient to learn good priors that would enable reliable completion of real-world measurements. Instead, our method is directly trained on noisy real-world observations, which yields the best results.

\subsection{Ablation Study}
\label{subsec:ablation}

\begin{table*}
\caption{\textbf{Effect of using geometric priors.} We report the point cloud completion performance of our method w.r.t the geometric priors used. Geometric priors assist in learning the object shapes while reducing the noise from the reconstructions. Using both cues in combination leads to the best performance.}
\label{tab:ablation1}
\centering
\makebox[\textwidth][c]
{\resizebox{1.0\linewidth}{!}{
    \begin{tabular}{l|ll|ll>{\columncolor[gray]{0.9}}ll|ll>{\columncolor[gray]{0.9}}ll|ll>{\columncolor[gray]{0.9}}ll}
    \toprule
    \multicolumn{1}{c}{\textbf{}}  & \multicolumn{2}{c|}{\textbf{}} & \multicolumn{4}{c|}{Chair} & \multicolumn{4}{c|}{Lamp} & \multicolumn{4}{c}{Table} \\
    \cmidrule{4-15} & $\mathcal{L}_{S}$ & $\mathcal{L}_{D}$ & rec.$\uparrow$  & prec.$\uparrow$ & F1$\uparrow$ & EMD$\downarrow$ & rec.$\uparrow$  & prec.$\uparrow$ & F1$\uparrow$ & EMD$\downarrow$ & rec.$\uparrow$  & prec.$\uparrow$ & F1$\uparrow$ & EMD$\downarrow$  \\
    \hline \hline
    A &  &  & 0.212 & 0.200 & 0.200 & 64.24 & 0.209 & 0.190 & 0.193 & 59.00 & 0.236 & 0.284 & 0.250 & 68.83   \\
    B  & \checkmark &   & 0.226 & 0.195 & 0.204 & 63.76 & 0.249 & 0.280 & 0.252 & \textbf{47.33} & 0.249 & 0.278 & 0.254 & 67.22 \\
    C   &  & \checkmark  & 0.230 & 0.201 & 0.209 & 63.39 & 0.247 & 0.287 & 0.253 & 48.20 & 0.251 & 0.287 & 0.259 & 67.80  \\
    D & \checkmark & \checkmark  & \textbf{0.232} & \textbf{0.211} & \textbf{0.215} & \textbf{61.29} & \textbf{0.254} & \textbf{0.300} & \textbf{0.264} & 47.58 & \textbf{0.256} & \textbf{0.300} & \textbf{0.268} & \textbf{66.15} \\ 
    \bottomrule
    \end{tabular}
}}
\end{table*}

\begin{figure*}[h]
\centering
  \includegraphics[width=0.90\linewidth]{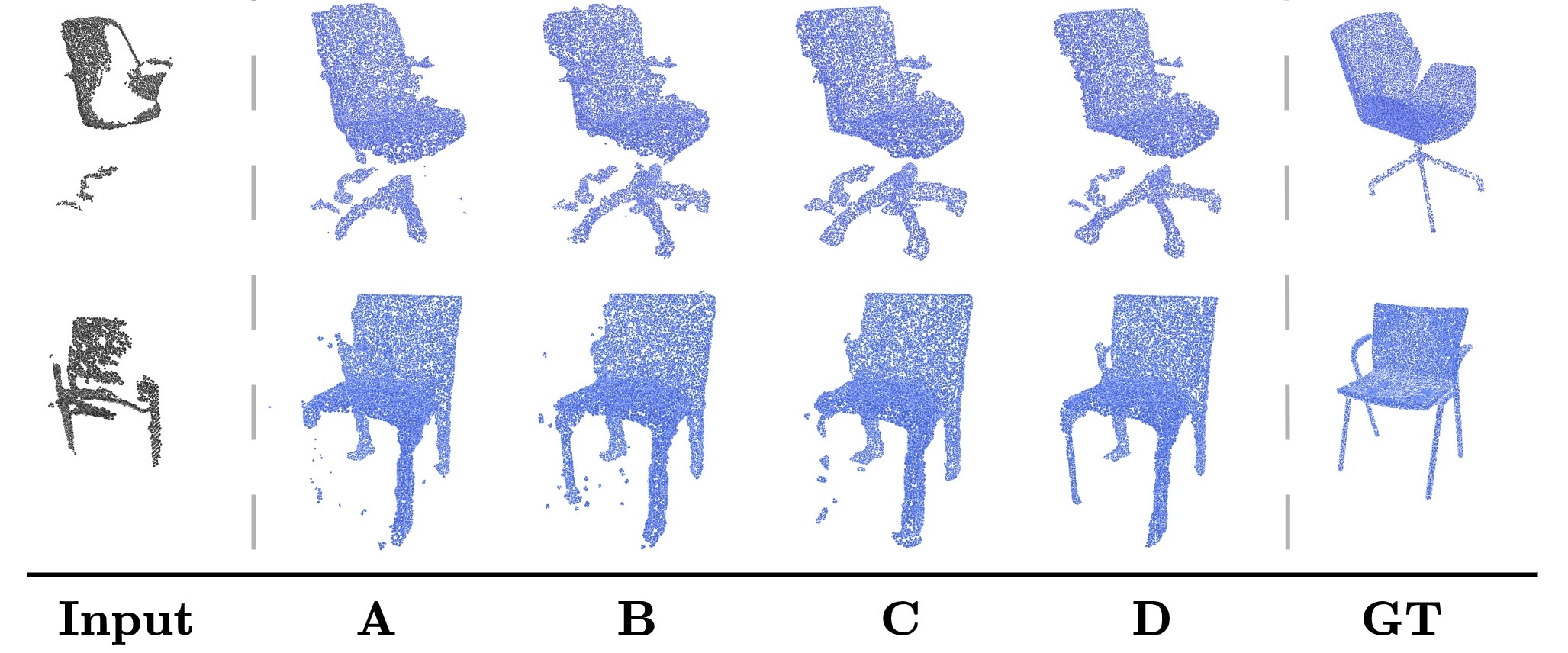}
   \caption{\textbf{Effect of using geometric priors.} Both of the geometric priors improve the reconstruction quality significantly, with the best performance being achieved when using both. With silhouette supervision, the predictions are more complete and have a better overall structure. With depth supervision, reconstructions become smoother with the elimination of noise.}
   \label{fig:ablation1qual}
\end{figure*}

\noindent\textbf{Effect of using geometric priors.} To investigate the effectiveness of each of the key components in our design, an ablation study is conducted on the ScanNet dataset. We train with four configurations: (\textbf{A}) raw setting of our model defined in Sec. \ref{subsec:subsec1} with only the reconstruction loss $\mathcal{L}_{t}$, (\textbf{B}) in addition to the raw setting, we add silhouette supervision, (\textbf{C}) we add depth supervision to the raw setting, (\textbf{D}) our full model which jointly learns shape information and smoothness prior by optimizing $\mathcal{L}$ defined in Sec. \ref{subsec:subsec2}. We report quantitative results in Table \ref{tab:ablation1} and present the visual comparison in Fig. \ref{fig:ablation1qual}. 

(\textbf{A}) without using depth or silhouette information already outperforms the baselines in terms of both F1-score and EMD on \textit{chair} and \textit{table} categories, and achieves comparable results on the \textit{lamp} category. Adding silhouettes (\textbf{B}) improves the recall, indicating more complete reconstructions. The inclusion of $\mathcal{L}_{S}$ helps the network in learning the overall object shapes more effectively. However, despite this improvement, the completion results still display noise and lack the desired smoothness, often characterized by wiggly shape boundaries and points appearing disconnected. From Fig. \ref{fig:ablation1qual}, it is shown that (\textbf{C}) reconstructs smoother surfaces particularly at the planar regions (\textit{e.g.} the back of the chair) and object boundaries, while also mitigating the noise observed in the visual results of (\textbf{A}) and (\textbf{B}). The two cues are complementary, with the best performance being achieved when combining them. Fig. \ref{fig:ablation1qual} shows that while (\textbf{D}) can complete the point clouds by recovering the overall shape of the objects, it can also reduce the reconstruction noise and produce sharper details.

\noindent\textbf{Effect of supervision completeness.} In this study, we investigate the relation between the shape completion performance and the number of views used for supervision. Point clouds from different numbers of views are combined to provide a more complete pseudo ground-truth to the network. Similar to the previous experiments, a single view is used as input. Table \ref{tab:ablation2} shows the results on \textit{chair} category of ScanNet. In general, as the ground-truth incompleteness increases, the ambiguity in how to complete the shape also increases owing to the multimodal nature of the completion task. However, in our experiments, we observe that the completion performance drops significantly as the number of views used to form the ground-truth increases. This indicates that when more noisy views are used for supervision, geometric priors cannot compensate for the accumulated noise coming from inaccurate real-world measurements anymore.

\begin{table}[h]
\caption{\textbf{Effect of supervision completeness.} We report the performance of our method w.r.t. number of views used to form the pseudo ground-truth. The completion performance drops significantly as the number of views used to form the ground-truth increases, indicating that geometric priors cannot compensate for the noise accumulated from the measurements anymore.}
\label{tab:ablation2}
\centering
    \begin{tabular}{l|ll>{\columncolor[gray]{0.9}}ll}
    \toprule
    \multicolumn{1}{c|}{Number of views}  &  \multicolumn{4}{c}{Chair}  \\
    & rec.$\uparrow$ & prec.$\uparrow$ & F1$\uparrow$ & EMD$\downarrow$  \\
    \hline \hline
    2-views & \textbf{0.232} & \textbf{0.211} & \textbf{0.215} & \textbf{61.29}  \\
    3-views & 0.205 & 0.163 & 0.177 & 68.29   \\
    4-views  & 0.207 & 0.159 & 0.175 & 69.50  \\
    5-views & 0.200 & 0.146 & 0.164 & 74.05  \\ 
    \bottomrule
    \end{tabular}
\end{table}

\begin{figure*}[h]
  \centering
   \includegraphics[width=0.9\linewidth]{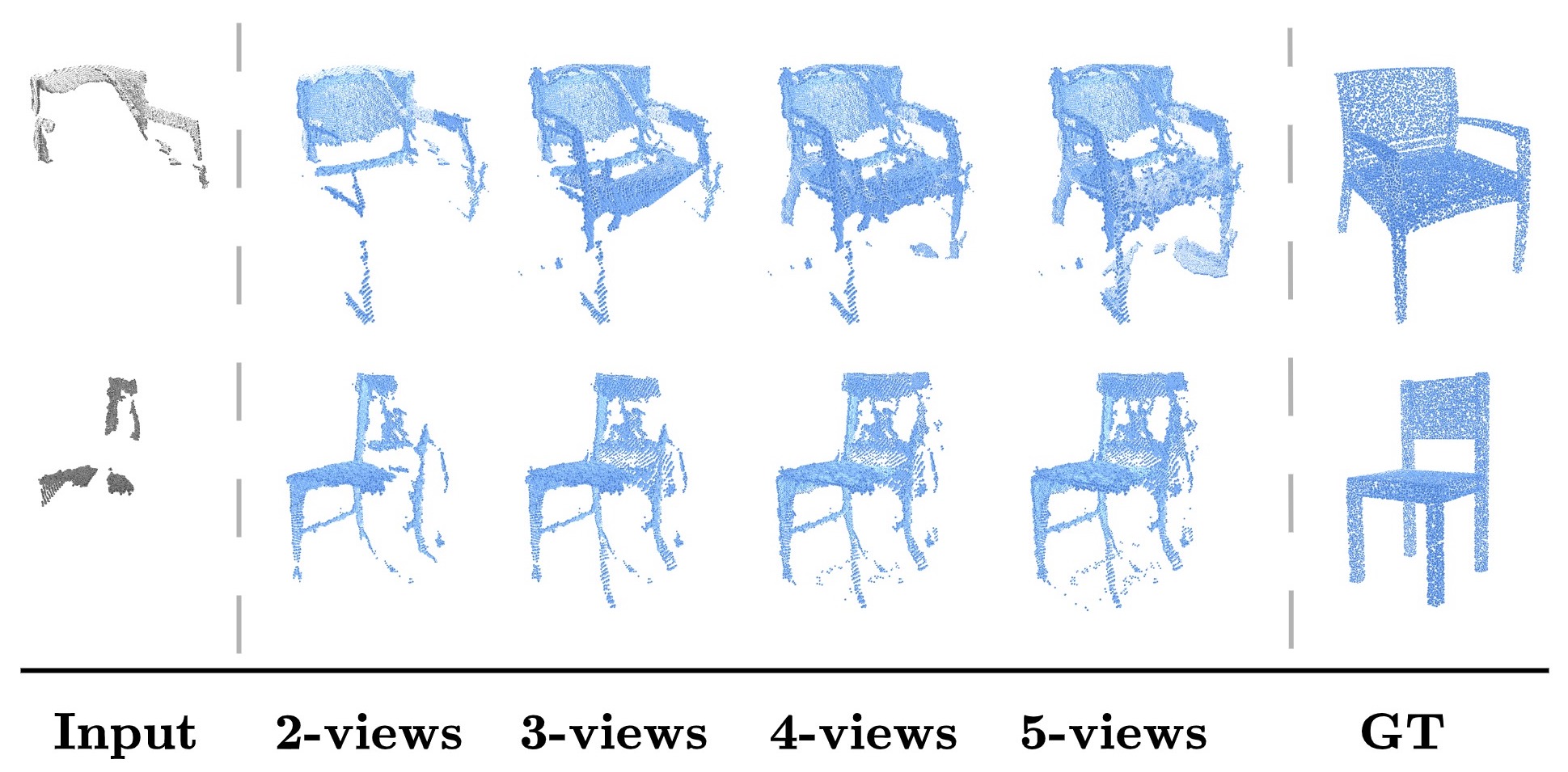}
   \caption{\textbf{Effect of supervision completeness.} Visual comparison of pseudo ground-truths generated by combining point clouds from different numbers of views. From left to right: input partial point cloud from a single view, pseudo ground-truths generated by combining point clouds from different numbers of views, and ground-truth CAD model alignments from Scan2CAD annotations. When more noisy views are combined for supervision, the noise from inaccurate real-world measurements accumulates, resulting in disconnected points, non-smooth surfaces, and thickened object parts due to non-overlapping point clouds from back-projected views.}
   \label{fig:supppseudogt}
\end{figure*}

We also provide a visual comparison of pseudo ground-truths generated by combining point clouds from different numbers of views in Fig. \ref{fig:supppseudogt}. When more views are combined for supervision, the noise from inaccurate real-world measurements accumulates, resulting in disconnected points, non-smooth surfaces and thickened object parts due to non-overlapping point clouds of back-projected views.

\noindent\textbf{Effect of inference strategy.} To investigate whether allowing modifications to the input at inference time enhances the generation performance, an ablation study is conducted on the \textit{chair} object category of ScanNet dataset. We perform the inference with two configurations: (\textbf{1}) modifications (M) are also allowed within the known regions during inference, and number of sampling steps is set to $T=1000$, (\textbf{2}) default setting for inference where we fix (F) the partial input and use $T=1000$ steps for the sampling. For both of the experiments (\textbf{1-2}), only the unoccupied input voxels are initialized with Gaussian noise while providing the input to the generation process.

We report quantitative results in Table \ref{tab:suppablation3}. When modifications are allowed on the partial input, the performance drops  as the reconstructed geometry deviates significantly from the ground-truth CAD model alignment.

\begin{table}[h]
\caption{ \textbf{Effect of inference strategy.} When modifications are allowed on the partial input, the performance drops as the reconstructed geometry deviates significantly from the ground-truth CAD model alignment.}
\centering
    \begin{tabular}{l|ll>{\columncolor[gray]{0.9}}ll}
    \toprule
    \multicolumn{1}{c|}{Inference strategy}  &  \multicolumn{4}{c}{Chair}  \\
    & rec.$\uparrow$  & prec.$\uparrow$ & F1$\uparrow$ & EMD$\downarrow$  \\
    \hline \hline
    (1) M, $T=1000$  & 0.096 & 0.097 & 0.100 & 112.28  \\
    (2) F, $T=1000$ & \textbf{0.228} & \textbf{0.186} & \textbf{0.200} & \textbf{64.36}  \\
    \bottomrule
    \end{tabular}
\label{tab:suppablation3}
\end{table}

\begin{table*}[h]
\caption{Complexity analysis. We report the number of parameters and computation cost (FLOPs) of all methods.}
\centering
\resizebox{0.6\linewidth}{!}{
    \begin{tabular}{l|ll}
    \toprule
    \multicolumn{1}{c|}{\textbf{Method}}  &  \multicolumn{2}{c}{}  \\
    & Params (M) & FLOPs (G) \\
    \hline \hline
    PoinTr \citep{pointr} & 42.50 & 17.92   \\
    Snowflake \citep{snowflakenet} & 19.32 & 20.78  \\
    AdaPoinTr \citep{adapointr} & 32.49 & 30.28   \\
    P2C \citep{p2c} & 112.03 & 13.72   \\
    MPC \citep{mcn} & 13.20 & 3.80  \\
    ShapeFormer \citep{shapeformer} & 324.95 & 139.70   \\
    \hline
    Ours & 78.53 & 114.28  \\
    \bottomrule
    \end{tabular}}
\label{tab:complexityanalysis}
\end{table*}

\subsection{Complexity Analysis}
\label{subsec:complexity analysis}

We report the number of parameters and computation cost (FLOPs) of all methods in Table \ref{tab:complexityanalysis}. Our method has a lower computational cost than \citep{shapeformer} which also employs a 3D architecture, showing our 3D U-Net's lightweight design.

\section{Conclusion}
\label{sec:conclusion}

This paper introduces an innovative self-supervised framework \textit{RealDiff}, designed for real-world point cloud completion. \textit{RealDiff} approaches the completion task as a conditional generation problem, simulating a diffusion process across the absent portions of objects while conditioning the generation on the  partial input observations. To improve shape learning and minimize reconstruction noise, our method leverages geometric cues, eliminating the need for synthetic priors. Extensive experiments demonstrate that the proposed approach consistently attains state-of-the-art performance in real-world point cloud completion.

\backmatter

\bmhead{Acknowledgements}

We thank members of the Bosch-UvA Delta Lab for helpful discussions and feedback. This project was generously supported by the Bosch Center for Artificial Intelligence.

\section*{Declarations}

\bmhead{Data Availability} This paper uses public datasets to conduct experiments. The datasets are available at the following URLs. ScanNet \citep{dai2017scannet}: \href{http://www.scan-net.org/}{http://www.scan-net.org/}. Redwood 3DScans \citep{redwood3d}: \href{http://redwood-data.org/3dscan/}{http://redwood-data.org/3dscan/}. ShapeNet \citep{chang2015shapenet}: \href{https://shapenet.org/}{https://shapenet.org/}.

\bigskip

\begin{appendices}

\section{Implementation Details}\label{secA1}

\textbf{Architecture Details.} \textit{RealDiff} is implemented on the PyTorch \citep{paszke2019pytorch} platform. We use a 3D U-Net \citep{3dunet} with 4 down and 3 up blocks as our main architecture. Diffusion time \textit{t} is specified by adding a sinusoidal positional embedding, as used in Transformers \citep{transformer}. The obtained temporal embedding is concatenated with the features before sending the input into the next resolution level. Each block consists of a convolutional block followed by group normalization \citep{groupnorm}, time embedding concatenation, and self-attention blocks. We use a voxel size of 2.5$cm$ with a grid of $64 \times 64 \times 64$ voxels.

\noindent \textbf{Training.} The network training is performed on one NVIDIA RTX A6000 GPU, using an Adam optimizer with batch size 16, and an initial learning rate of 0.0001. We initially train our network without the rendering losses for 250 epochs. During training, we input $\boldsymbol{\mathrm{x}}_{t} = (\boldsymbol{\mathrm{\tilde{x}}}_{t}, \boldsymbol{\mathrm{c}}_{0})$ into $f_{\theta}$ to predict the complete shape $\boldsymbol{\mathrm{{x}}}_{gt}$. To keep $\boldsymbol{\mathrm{c}}_{0}$ fixed, we compute the loss only for the regions of $\boldsymbol{\mathrm{\tilde{x}}}_{t}$ by masking out the regions of $\boldsymbol{\mathrm{c}}_{0}$ from the model output. Then, we continue training with the auxiliary geometric losses until convergence. While training with geometric priors, modifications are allowed at the previously fixed voxels by computing the loss also for the regions of $\boldsymbol{\mathrm{c}}_{0}$ within the output. With this approach, it is ensured that while the diffusion model initially learns to complete shapes by conditioning on  $\boldsymbol{\mathrm{c}}_{0}$ in the first phase of training, it also acquires the ability to remove noise from the reconstructions in the second phase.
\\

\noindent \textbf{Hyperparameters.} Following \cite{diffmodelbase2}, we define the variance schedule $\beta_{t}$ according to a linear schedule. We set $\beta_{0} = 10^{-4}$ and $\beta_{T} = 2 \times 10^{-2}$, and linearly interpolate other $\beta_{t}$'s. We set the number of time steps $T$ in the diffusion process to 1000 in all the experiments.

\section{Additional Dataset Details}\label{secA4}

\textbf{ScanNet.} ScanNet \citep{dai2017scannet} projects 3D meshes annotated by crowd workers into 2D frames according to the camera trajectory of the sequence. Despite the existence of per-frame 2D instance masks, the instance IDs are not consistent across different frames for a particular object. Hence, in order to ensure consistency across all frames, we re-labeled the 2D instance masks using the 3D instance masks from the externally annotated meshes. More specifically, we back-project a 2D instance mask to see if it is overlapping with any of the 3D instance masks and assign the instance ID accordingly in the case of a match. We utilize the 2D instance masks as the externally obtained silhouettes for the objects by excluding other instances from the frame. 

To evaluate point cloud completion on real-world observations, the Scan2CAD \citep{scan2cad} annotations are used of CAD model alignments from ShapeNet to the extracted objects. For each CAD model, 16384 points are sampled for evaluation. To obtain the predicted point cloud, first a mesh is extracted from the predicted voxel grid via the Marching Cubes algorithm, and then 16384 points are sampled from the surface of the mesh.

\noindent\textbf{ShapeNet.} ShapeNet dataset \citep{chang2015shapenet} is used to evaluate the effectiveness of synthetic shape priors for real-world performance. For each object, we first randomly sample 16384 points from the surface to obtain a ground-truth point cloud for that object. To train the baselines on the ShapeNet dataset, the data pre-processing protocol introduced by \cite{pointr} is taken to obtain input partial observations. More specifically, following PoinTr \citep{pointr}, we first select a viewpoint and then remove $n$ furthest points from the viewpoint to obtain a partial point cloud. $n$ is chosen as 8192. During evaluation, we use randomly sampled 8192 points from the real-world input data of ScanNet and predict the output with 16384 points. We also sample 16384 points from the CAD model.

\section{Details of the Evaluation Metrics}\label{secA2}

\subsection{{Metrics for Single Completion Performance}}

\textbf{F1-score ($\uparrow$).} We adopt F1-score as our main evaluation metric to explicitly evaluate the distance between object surfaces as suggested by \cite{maximf1}. It is defined as the harmonic mean between precision and recall, which measure \textit{accuracy} and \textit{completeness} of the reconstructions respectively. Consider $\boldsymbol{\mathrm{\hat{P}}}$ and $\boldsymbol{\mathrm{P}}$ as the predicted and ground-truth point clouds, with ${\hat{p}} \in \boldsymbol{\mathrm{\hat{P}}}$ and ${p} \in \boldsymbol{\mathrm{P}}$, then the precision, recall and F1-score are defined as follows:
\begin{equation}
    \mathcal{L}_{P}(\tau) = \text{mean}_{{\hat{p}}\epsilon \boldsymbol{\mathrm{\hat{P}}}}(\min_{{p}\epsilon \boldsymbol{\mathrm{P}}}\left\|{\hat{p}}-{p}\right\| < \tau)
\end{equation}

\begin{equation}
    \mathcal{L}_{R}(\tau) = \text{mean}_{{p}\epsilon \boldsymbol{\mathrm{P}}}(\min_{{\hat{p}}\epsilon \boldsymbol{\mathrm{\hat{P}}}}\left\|{\hat{p}}-{p}\right\| < \tau)
\end{equation}

\begin{equation}
    F_{1} = \frac{2 \cdot \mathcal{L}_{P}(\tau) \cdot \mathcal{L}_{R}(\tau)}{\mathcal{L}_{P}(\tau) + \mathcal{L}_{R}(\tau)}
\end{equation}
\noindent where $\tau = 10^{-2}$ is the pre-defined distance threshold.

\noindent\textbf{Earth Mover's Distance ($\downarrow$).} Following prior work \citep{diffpc1, diffpcextra1, ganinversion}, we also employ Earth Mover's Distance (EMD) which attempts to find the optimized solution of transforming one point set to another. Consider $\boldsymbol{\mathrm{\hat{P}}}$ and $\boldsymbol{\mathrm{P}}$ as the predicted and ground-truth point clouds of equal size, their EMD is defined by: 

\begin{equation}
     d_{EMD}(\boldsymbol{\mathrm{\hat{P}}}, \boldsymbol{\mathrm{P}}) = \min_{\phi: \boldsymbol{\mathrm{\hat{P}}} \to \boldsymbol{\mathrm{P}}} \sum_{\hat{p} \in \boldsymbol{\mathrm{\hat{P}}}}^{} \left\| \hat{p} - \phi (\hat{p}) \right\|_{2}
\end{equation}
\noindent where $\phi$ is a bijection. Similar to \cite{ganinversion, liu2020morphing}, EMD is used to quantify how uniform the density of the predicted point cloud is (we refer to this property as \textit{uniformity}). We use the implementation provided by PoinTr \citep{pointr}. For the probabilistic methods, random runs are used for evaluations.

\subsection{{Metrics for Multi-modal Completion Performance}}

Our multi-modal completion performance is evaluated using Minimal Matching Distance (MMD) capturing the completion quality, Total Mutual Difference (TMD) measuring the completion diversity, and Unidirectional Hausdorff Distance (UHD) focusing on the completion fidelity, as suggested by \cite{mcn}. 

Consider $\mathcal{T}_{p}$ and $\mathcal{T}_{c}$ as the sets of partial point clouds and ground-truth complete point clouds. $\mathcal{T}_{c}$ consists of CAD model alignments in our case. $k$ completed shapes $c_{ij}, \: j=1..k$, are generated for each partial shape $p_{i}$ in $\mathcal{T}_{p}$. $\mathcal{G}_{c}=\{c_{ij}\}$ indicates the set of generated shapes, and $k=10$ in our evaluations. 

\noindent\textbf{Minimal Matching Distance  ($\uparrow$).} Different from \cite{mcn}, for each shape $s_{i}$ in $\mathcal{T}_{c}$, its nearest neighbor $\mathbf{N}(s_{i})$ is found in $\mathcal{G}_{c}$ by using F1-score as the metric by selecting the completed shape ${c_{ij}}$ with the highest F1-score. MMD is then defined as follows:

\begin{equation}
      \text{MMD} =  \frac{1}{|\mathcal{T}_{c}|} \sum_{s_{i} \in \mathcal{T}_{c}}^{} F_{1}(s_{i}, \mathbf{N}(s_{i})) 
\label{eq:mmd}
\end{equation}


We report an upper bound performance using the main evaluation metrics (used in Tab. 1 and Tab. 3-5 of the main paper) for all methods in Appendix \ref{secA3} based on the  selected nearest neighbor shapes. Therefore, F1-score is preferred for the nearest neighbor selection to maintain consistency with the main evaluation metrics. For the rest of the multi-modal evaluation metrics, we use the original definitions.


\noindent\textbf{Total Mutual Difference  ($\uparrow$).} For each of the completed shapes $c_{ij} (1 \le j \le k)$, from the same partial shape $p_{i}$, an average Chamfer distance is calculated to the other $k-1$ completed shapes. Then, the resulting $k$ distances are summed up. The Total Mutual Difference (TMD) is then computed as the average value across partial shapes in $\mathcal{T}_{p}$ as follows:

\begin{equation}
\begin{split}
     \text{TMD} = \frac{1}{|\mathcal{T}_{p}|}\sum_{i=1}^{|\mathcal{T}_{p}|}(\sum_{j=1}^{k}\frac{1}{k-1}\sum_{1\le l \le k, l\neq j}^{} d^{\text{CD}}(c_{ij}, c_{il})) \\ = \frac{1}{|\mathcal{T}_{p}|}\sum_{i=1}^{|\mathcal{T}_{p}|}(\frac{2}{k-1} \sum_{j=1}^{j} \sum_{l=j+1}^{k} d^{\text{CD}}(c_{ij}, c_{il}) )
\end{split}
\end{equation}

\noindent\textbf{Unidirectional Hausdorff Distance  ($\downarrow$).} The average Unidirectional Hausdorff Distance (UHD) is computed from the partial shape $p_{i}$ in $\mathcal{T}_{p}$ to each of its completed shapes $c_{ij}, \: j=1..k$, where $(1 \le j \le k)$:

\begin{equation}
\text{UHD} = \frac{1}{|\mathcal{T}_{p}|}\sum_{i=1}^{|\mathcal{T}_{p}|}(\frac{1}{k}\sum_{j=1}^{k}d^{\text{HD}}(p_{i},c_{ij})),
\end{equation}

\noindent where $d^{\text{HD}}$ denotes the unidirectional Hausdorff distance.

\section{Upper Bound Performance of Multi-modal Methods}\label{secA3}

\begin{table*}[h]
\caption{Minimal matching distance evaluation on the ScanNet dataset \citep{dai2017scannet} at the resolution of 16384 points. The reported F1 score is computed based on \ref{eq:mmd}. Recall, precision and EMD are reported based on the nearest neighbor shapes. Our method consistently outperforms the baseline methods, showing the stability of our approach.}
\centering
\makebox[\textwidth][c]
{
    \resizebox{1.0\linewidth}{!}{
    \begin{tabular}{l|ll>{\columncolor[gray]{0.9}}ll|ll>{\columncolor[gray]{0.9}}ll|ll>{\columncolor[gray]{0.9}}ll}
    \toprule
    \multicolumn{1}{c|}{\textbf{Method}}  &  \multicolumn{4}{c|}{Chair} & \multicolumn{4}{c|}{Lamp} & \multicolumn{4}{c}{Table} \\
    
    \cmidrule{2-13} & rec.$\uparrow$   & prec.$\uparrow$  & F1$\uparrow$  & EMD$\downarrow$
    & rec.$\uparrow$ & prec.$\uparrow$  & F1$\uparrow$  & EMD$\downarrow$ 
    & rec.$\uparrow$   & prec.$\uparrow$ & F1$\uparrow$ & EMD$\downarrow$  \\
    \hline \hline
    
    MPC \citep{mcn}  & 0.11 & 0.12 & 0.11 & 85.67 &  0.11 & 0.10 & 0.10 & 84.12 & 0.17 & 0.17 & 0.17 & 93.30 \\
    ShapeFormer \citep{shapeformer}  & 0.19 & \textbf{0.22} & 0.20 & \textbf{57.76} &  0.15 & 0.14 & 0.15 & 70.80 & 0.16 & 0.16 & 0.16 & 134.55 \\
    \hline 
    Ours & \textbf{0.24} & \textbf{0.22} & \textbf{0.23} & 58.61 &  \textbf{0.25} & \textbf{0.30} & \textbf{0.26} & \textbf{47.34} & \textbf{0.26} & \textbf{0.31} & \textbf{0.28} & \textbf{65.32} \\
    \bottomrule
    \end{tabular}}
}
\label{tab:supprealworldmmd}
\end{table*}

We further report an upper bound completion performance of multi-modal methods utilizing the MMD metric. MMD is computed between the set of partial shapes and ground-truth complete shapes as described in Appendix \ref{secA2}. While the nearest neighbor shape selection is performed based on the highest F1-score, we additionally compute precision, recall and EMD based on the selected shapes. The quantitative results are presented in Table \ref{tab:supprealworldmmd}. Note that only MPC \citep{mcn}, ShapeFormer \citep{shapeformer} and our method are able to produce multiple plausible completions from a single input, and therefore we only provide the results for them. PoinTr \citep{pointr}, SnowflakeNet \citep{snowflakenet}, AdaPoinTr \citep{adapointr} and P2C \citep{p2c} generate the same completions at each run. 

\textit{RealDiff} consistently outperforms the baseline methods, showing the stability of our approach. The F1-score of ShapeFormer improves relatively more than MPC and our method in \textit{lamp} and \textit{table} categories. This is due to deviations among completions from ShapeFormer. While it generates reasonable geometries in some of the runs, it struggles to consistently produce accurate reconstructions.

\end{appendices}


\bibliography{sn-bibliography}

\end{document}